\documentclass[11pt]{article}
\usepackage{acl2015}
\usepackage{times}
\usepackage{url}
\usepackage{latexsym}
\usepackage{graphicx}
\usepackage{amsmath}
\usepackage[toc,page]{appendix}
\usepackage{caption}
\usepackage[colorlinks=true,citecolor=blue,backref=page]{hyperref}
\usepackage{subcaption} %For subtables and subfigures
\usepackage[justification=centering]{caption} %This centers captions in figures
\usepackage{chngcntr} %This package is used for the counterwithin commands
\usepackage{fancyhdr}%This package controls header and footer latyout
\usepackage[table]{xcolor}  %For coloring rows in tables
\usepackage{multirow}
\usepackage{tabularx} %For automatically adjusting column width
\usepackage{tabulary,booktabs} %Similar
\usepackage{ragged2e} %for making tables prettier
\usepackage{changepage} %For indenting paragraphs
\usepackage{multicol}
\usepackage[htt]{hyphenat}
\usepackage{lipsum}
\usepackage{caption}
\usepackage{twemojis}
\usepackage{array}
\newcolumntype{P}[1]{>{\raggedright\arraybackslash}p{#1}}
\usepackage{apacite} %For citing APAstyle

\title{Profiling Irony \& Stereotype: \\ Exploring Sentiment, Topic, and Lexical Features}

\author{T. L. R. Krols\thanks{\hspace{1mm} Equal Contribution} \\
  University of Copenhagen \\
  {mdl163@alumni.ku.dk} \\\And
  Marie Mortensen* \\
  University of Copenhagen\\
  {hgt160@alumni.ku.dk} \\\And
  Ninell Oldenburg*\\
  University of Copenhagen\\
  {gxf740@alumni.ku.dkn} \\}

\date{}

\begin{document}
\maketitle
\begin{abstract}
Social media has become a very popular source of information. With this popularity comes an interest in systems that can classify the information produced. This study tries to create such a system detecting irony in Twitter users. Recent work emphasize the importance of lexical features, sentiment features and the contrast herein along with TF-IDF and topic models. Based on a thorough feature selection process, the resulting model contains specific sub-features from these areas. Our model reaches an F1-score of 0.84, which is above the baseline. We find that lexical features, especially TF-IDF, contribute the most to our models while sentiment and topic modeling features contribute less to overall performance. Lastly, we highlight multiple interesting and important paths for further exploration.
\end{abstract}
%This text contains \charactercount{main} characters.

\section{Introduction}\label{sec:introduction}
Social media allows people to share online content easy and fast. As this content is not necessarily only positive, the voices for regulation became louder within the last years. Such regulation contains subtasks as the identification of misinformation, fake news, and not at least stereotypes against minority groups. One famous example of such regulation is the ban of the American president from the social media platform Twitter in 2021 partly due to his spreading of stereotypes \cite{twitter_permanent_2021}.

Since irony as well as stereotyping have been proven to have a persuasive meaning \shortcite{brown_prejudice_2011, ghosh_semeval-2015_2015} and an influence on how we perceive reality such as political debates and influence the decisions we make \shortcite{shu_fake_2017}, the automatic identification of irony and stereotyped language is becoming a central point of interest. The 2022 PAN shared task IROSTEREO on \textit{Profiling Irony and Stereotype Spreaders} on Twitter\footnote{See task under \url{https://pan.webis.de/clef22/pan22-web/author-profiling.html}} tries to investigate this issue and enhance research on it. The primary goal of the task is to identify and distinguish between ironic authors based on their tweets even though stereotype identification is stated as subtask among those that have been identified as ironic \shortcite{ortega-bueno_pan_2022}. We will give a brief introduction into the field of automatic irony detection in Section \ref{sec:related}, motivate our approach of doing an exploratory study in which we are trying to identify and relate the influence of sentiment, topic, and lexical features on irony detection in Section \ref{sec:methods}. Lastly, we present, discuss, and relate our findings to the current state-of-the-art (SOTA) models in Sections \ref{sec:results} and \ref{sec:discussion}.

\section{Related Work}\label{sec:related}
Linguistically, irony is seen as a communicative act of implicitly expressing the opposite of what is literally said \shortcite{haverkate_speech_1990, wilson_verbal_1992, utsumi_verbal_2000}. For example, the utterance \textit{I love the pouring rain} would be identified as ironic with the contrasting elements \textit{love} and \textit{pouring rain}. The identification of such is described as non-trivial even to human subjects due to its dependence on variables like context, intonation, and speaker \shortcite{giora_irony_1995,barbe_irony_1995,utsumi_verbal_2000}. Such conditions make an automatic, text-based identification even more difficult. The existing attempts are very heterogeneous but can be broadly divided into either Machine Learning (ML)-based approaches with hand-crafted features or Deep Learning (DL)-based \shortcite{joshi_automatic_2016}.

We want to emphasize that irony and sarcasm are often mentioned in the same context \shortcite{devlin_bert_2019, potamias_transformer-based_2020}, as they are described as having a shared contrastive meaning and expression as described above \shortcite{attardo_multimodal_2003}, which is why we will also include studies on sarcasm detection into our analysis.

\subsection{Machine Learning Approaches}

One of the first approaches on ML irony detection \shortcite{carvalho_clues_2009} found that textual representations of oral and gestural cues like emoticons, punctuation or expressions of laughter are helpful for identifying ironic sentences in Portuguese online newspaper. \shortciteA{bratulic_dots_2021} further embrace the importance of surface patterns for different tested and mostly neural network models. Exploiting syntactic and lexical features also seems promising both from a linguistic \cite{kreuz_two_1995} as well as from a probabilistic standpoint \shortcite{kreuz_lexical_2007, riloff_sarcasm_2013}. The most prominent morphosyntactic features among the area of irony detection are word or character N-grams \shortcite{davidov_semi-supervised_2010, reyes_humor_2012, reyes_multidimensional_2013, wu_thu_ngn_2018}, tweet length \shortcite{hernandez-farias_applying_2015}, sentiment-conveying features like emoticons \shortcite{hernandez-farias_applying_2015}, or sentiment-lexicon implementations \shortcite{reyes_humor_2012, liebrecht_cc_perfect_2013, joshi_automatic_2016}. Another seemingly important trait to extract from texts is a term frequency–inverse document frequency (TF-IDF) which falls in between the field of lexical and topic features \shortcite{khalifa_ensemble_2019}. It will be considered as a lexical feature here.

However, other successful ML approaches focus more on the explicit or implicit clash of sentiment \shortcite{riloff_sarcasm_2013, farias_irony_2016, potamias_transformer-based_2020} as described in \shortciteA{utsumi_verbal_2000}. These are sometimes paired with word embeddings and syntactic features as in \shortciteA{barbieri_modelling_2014, wu_thu_ngn_2018, mohammad_semeval-2018_2018}. For instance, \shortciteA{mohammad_semeval-2018_2018}, obtain those sentiment features via the AffectiveTweet package \shortcite{bravo-marquez_affectivetweets_2019} whereas \shortciteA{barbieri_modelling_2014} measure the level of contradiction using the American National Corpus Frequency Data \shortcite{fillmore_american_1999} as a source.

Furthermore, an understudied yet promising approach (F1-score of up to 84.4\%) includes topic modeling \shortcite{khalifa_ensemble_2019}. Also \shortciteA{nozza_unsupervised_2016} achieve good results with their unsupervised topic modeling approach when compared to supervised irony detection models of that time.

Lastly, the predominant classifiers used are Naïve Bayes (NB) \shortcite{reyes_humor_2012, farias_irony_2016, chia_machine_2021}, Support Vector Machines (SVM) \shortcite{carvalho_clues_2009, farias_irony_2016, chia_machine_2021}, or tree based classifiers like Random Forest (RF) \shortcite{van_hee_semeval-2018_2018, barbieri_modelling_2014} and Decision Trees (DT) \shortcite{reyes_multidimensional_2013, farias_irony_2016, barbieri_modelling_2014}. According to a comparison of different classifiers by \shortciteA{buschmeier_impact_2014}, Logistic Regression (LR), DT and SVM perform best.

\subsection{Deep Learning Approaches}

Despite the successes of ML approaches, the time-consuming aspect of using a list of engineered features has been criticized \shortcite{potamias_transformer-based_2020} and the rise of fast and efficient DL models across a wide range of applications has also reached the area of irony detection.

Early DL approaches were e.g. undertaken by \shortciteA{amir_modelling_2016} improving their baseline of more than 2 percentage points by using user embeddings (mostly learned text surface features specific to a user) or \shortciteA{hazarika_cascade_2018} successfully pairing those user embeddings with feature extractors and raising SOTA by up to 8\% depending on the dataset and data cleaning. Both used Convolutional Neural Networks (CNN).

\begin{figure*}[!ht]
    \centering
    \includegraphics[width=16cm]{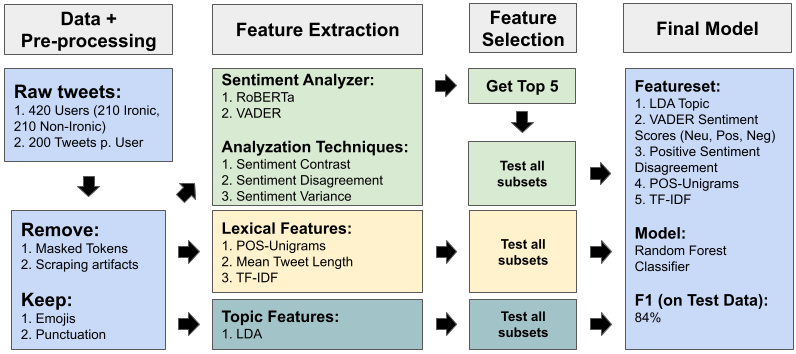}
    \caption{Pipeline of the methodology we applied to come from the provided data to an irony prediction model.}
    \label{fig:pipeline}
\end{figure*}

Afterwards, during the SemEval-2018 shared task 3 on irony detection in English tweets \shortcite{van_hee_semeval-2018_2018}, the winning team, \shortciteA{wu_thu_ngn_2018}, achieved an F1-score of up to 70.54\% with densely connected LSTMs capturing hidden text representations. \shortciteA{van_hee_we_2018} also found that including implicit sentiment (or context) and capturing the contrast to explicit sentiment is highly successful. They describe implicit sentiment as the sentiment of a situation, such as \textit{going to the dentist}. \shortciteA{zhang_irony_2019} focused on that contrast, too, and achieved an 82.4\% F1-score with a Bidirectional LSTM (Bi-LSTM). Lastly, \shortciteA{ahuja_transformer-based_2021} found that accuracy highly depends on the used model comparing five ML models, six DL models and six transformer-based models resulting in a maximum F1-score of 88\% with their CNN.

While DL systems seem very promising, they are both very data-hungry and suffer from the problem of model interpretability \shortcite{ortega-bueno_multi-view_2022}. Even though they give very good predictions, the reasons for their excellent performance remains partly unknown.

%The applications for such a system range widely from opinion mining to detection of online harassment, sentiment analysis and author identification \cite{van_hee_semeval-2018_2018,chia_machine_2021}.

\section{Data}\label{sec:data}

The dataset consists of 200 English tweets from 210 ironic ("I") users and 210 non-ironic ("NI") user. The ground truth label "I" reflects that a certain number of tweets of the respective author is labeled as being ironic. The authors of the shared task place a special emphasis on Twitter authors, that convey stereotypes against women or the LGBT community \shortcite{ortega-bueno_pan_2022}. In contrast, Twitter authors labeled as "NI" do not reach that number of posts with ironic content.  The number of that critical mass of tweets is to the best of our knowledge unknown. To be clear, only the authors are labeled as ironic/non-ironic and not individual tweets.

Hashtags, links and mentions are all masked as either \textit{\#HASHTAG\#}, \textit{\#URL\#}, or \textit{\#USER\#}. However, emojis and emoticons are maintained as is in the original tweets. The complete dataset can be requested from the PAN task website\footnote{\url{https://pan.webis.de/clef22/pan22-web/author-profiling.html}} \shortcite{ortega-bueno_pan_2022}.

% tweets database x
% x tweets per authors x
% x authors x
% each author is labeled as ironic or onironic. (no labeled tweets as ironic) x
% ground truth labels obtained by....(how many people???) x
% unknown how many tweets make an author ironic.   x
% balanced dataset x
% hashtags + mentions marked out.  x
% emojis maintained x

\section{Methods}\label{sec:methods}
The overall aim of the task is to develop a model for the classification of users into ironic or non-ironic, following the intuition that a label, even by human judges, could be distinguished based on the history of other tweets of one author. 
The methodology applied to achieve this, which includes both data cleaning, feature engineering and model building, is illustrated in Figure \ref{fig:pipeline}.

\subsection{Pre-processing}
\begin{figure}[!ht]
    \centering
    \includegraphics{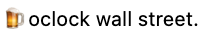}
    \caption{Example of processed tweet.}
    \label{fig:tweet}
\end{figure}

Before developing a baseline along with our suggested model, the data is subject to multiple processing steps. Initially all \texttt{.xml} files that each contain a user's 200 tweets are collected and combined. All masked tokens (\textit{\#HASHTAG\#}, \textit{\#URL\#}, or \textit{\#USER\#}) were removed as we focus our analysis on other linguistic aspects of the tweets. Scraping artifacts such as 'A\&amp;M' and other long unidentifiable sequences of characters and digits above a length of 30 are removed as well. An example of a processed tweet can be seen in Figure \ref{fig:tweet}. 

We made a train-test split of the Twitter users in the data where 70\% of the users were utilized for training our models, and 30\% was used for testing the final model's performance. 
\subsection{Feature Extraction}
We create some features on user level, whereas other features were created on tweet level in order to capture more detailed variations. Practically this means that some features are just one number per user (mean tweet length, dominant topic, cluster, and sentiment variation) whereas the other features result in a 200-dimensional vector per user as there is information per tweet. The code used for feature generation can be found on GitHub\footnote{\url{https://github.com/Tibor-Krols/LangProc3000}}.

\subsubsection{Sentiment Features}
Following the early analysis by \shortciteA{haverkate_speech_1990, wilson_verbal_1992} and \shortciteA{utsumi_verbal_2000} as well as more recent research on automatic irony detection via sentiment \shortcite{riloff_sarcasm_2013, farias_irony_2016, potamias_transformer-based_2020}, we exploit several sentiment features aiming to capture different aspects of irony, namely the sentiment contrast, disagreement and variance.

In general, sentiment seems to be an important aspect of irony as described in Section \ref{sec:related}. We decided to perform a sentiment analysis on each tweet with the sentiment analysis tool VADER (\textbf{V}alence \textbf{A}ware \textbf{D}ictionary for s\textbf{E}ntiment \textbf{R}easoning \shortcite{zainudin_sentiment_2015, hutto_vader_2014}. It gives rise to a positive, negative and neutral sentiment score each ranging from 0 to 1 and a so-called compound sentiment score for the overall text sentiment ranging from -1 to 1. On the other hand, we also obtain sentiment scores with the \textbf{R}obustly \textbf{O}ptimzed \textbf{BERT} \textbf{A}pproach RoBERTa \shortcite{liu_roberta_2019}, also resulting in a positive, negative and neutral sentiment score ranging from 0 to 1. The main difference between the two is that VADER calculates its scores based on lexicon and rules, whereas RoBERTa as a transformer model is fine-tuned on our tweet dataset and generates the scores based on more context.

\textbf{Sentiment Contrast.} Especially the clash of explicit or implicit sentiment aspect of irony seemed to have proven good results in both ML \shortcite{riloff_sarcasm_2013, farias_irony_2016, potamias_transformer-based_2020} as well as DL approaches \shortcite{van_hee_we_2018, zhang_irony_2019}. These are defined as either one sub-sentence or two bordering parts reflecting the opposite sentiment. We attempted to capture this contrast by calculating the trigram-based sentiment scores and subtracting the minimum sentiment from the maximum sentiment per tweet with RoBERTa. This approach follows the procedure in \citeA{rajadesingan_sarcasm_2015} where the contrast is defined as:

$$\Delta_{\text {sentiment }}=\max (S)-\min (S)$$

where $S$ are the trigram sentiment scores in a tweet.

\textbf{Sentiment variation.}
Furthermore, as an extension on sentiment contrast, we look at the the variation of sentiment contrast within a user. We do that by determining the standard deviation of both positive and negative sentiment contrast, again using trigrams and RoBERTa. Along with this measure, we also calculate the standard deviation of all positive, negative and neutral scores in a user's collection of tweet. 

\textbf{Sentiment Disagreement.} Lastly, we're investigating the aspect of irony that is being described as the difficulty of determining the true sentiment of the sentence, even for human annotators \shortcite{wilson_verbal_1992, giora_irony_1995, barbe_irony_1995, utsumi_verbal_2000}. We attempted to capture this aspect with a sentiment disagreement feature. The sentiment disagreement exploits the different structure and algorithmic properties both of our sentiment analyzers by calculating the squared distance between the standardized scores for positive, negative and neutral sentiments of each of the classifiers. This is a method usually used to clean data when observing different scores for different analyzers \shortcite{ghasiya_investigating_2021}. We hypothesized a low squared distance for easy-to-predict sentiment and in contrast a rather higher score for harder-to-predict sentiment.

\subsubsection{Topic Features}

To the best of our knowledge, topic analysis is an understudied phenomenon in irony detection. However, we hypothesize that the topic that authors talk about can be of relevance.

\textbf{LDA.} Following \shortciteA{nozza_unsupervised_2016}, we obtained a topic model with a Latent Dirichlet Allocation (LDA) \cite{jelodar_latent_2019} using the package \texttt{sklearn. decomposition.LatentDirichletAllocation} \shortcite{pedregosa_scikit-learn_2011}. LDA is a topic modeling technique that has as basic idea that every word $(w_n | w \in D_i)$ is subject to a topic distribution $T_m(w)$ and every document $D$ a distribution over topics $D_i(T)$, with $m=$ number of topics. In our model, $D_i =$ tweet$_n$ for $n\in 200$ tweets$\times 420$ users. The ideal number of topics was tested with perplexity resulting in \texttt{n\_components=5} at a score of 85.33. This suggests that all of the documents can be captured by five topic groups.

\textbf{Cluster.} Along with LDA, we cluster the TF-IDF vectors into five clusters using the \texttt{sklearn.cluster.KMeans} \cite{pedregosa_scikit-learn_2011}.

\subsubsection{Lexical Features}
Following the literature presented in Section \ref{sec:related} (e.g. \shortciteA{kreuz_lexical_2007, riloff_sarcasm_2013}), we analyzed different lexical features. These were Part-of-speech (POS) unigrams (e.g. \shortcite{buschmeier_impact_2014} calculated via the Natural Language Tool Kit (NLTK) \shortcite{bird_natural_2009}, and the mean tweet length per author \shortcite{hernandez-farias_applying_2015}.

\textbf{TF-IDF.} Further following the successes of the literature (e.g. \shortciteA{khalifa_ensemble_2019}), we calculated the term frequency - inverse document frequency (TF-IDF) per author via the \texttt{sklearn.feature\_extraction.text.TfidfVectorizer} \shortcite{pedregosa_scikit-learn_2011}. This makes it a feature on user level. We discarded all words that are in more than 95\% and less than 5\% of the tweets resulting in two word clusters of words that are labeled ironic or non-ironic, respectively. The most frequent words and bi-grams per label can be seen in Table \ref{tab:most_freq_words}.

\begin{table}[]
    \centering
\begin{tabular}{l|P{2.6cm}|P{2.6cm}}
            & Ironic & Not Ironic \\ \hline
Uni    & calm, guard, ukraine, re, think, war, im, know, yes, didnt 
            & women, gay, military, men, dont, need, it, you, black, im \\ \hline
Bi     & calm down, added closet, listing added, check listing, you re, white nationalist, human rights, feel like, looks like, things like
            & women sports, trans women, men women, you re, black people, trans people, high school, good thing, dont believe, im sorry \\                                         
\end{tabular}
    \caption{Listing of the 10 most frequent uni- and bigrams after we performed a TF-IDF analysis.}
    \label{tab:most_freq_words}
\end{table}

\subsection{Feature Selection}
Our process for feature selection is twofold.\\
\indent Firstly, based on related work in Section \ref{sec:related}, we determined the most important feature categories to be sentiment features, topic features and lexical features.\\
\indent Secondly, we analyzed which subset of features performs the best per feature category. For feature selection we measure performance based on F1-score over 5-fold cross validation (\texttt{sklearn.model\_selection.KFold} \shortcite{pedregosa_scikit-learn_2011}) with a RF classifier. This procedure is based on the methods presented in \citeA{kuhn_feature_2019}.
Since we only had few features for the categories of topic and lexical features, we tested all possible combinations. Because we have 14 sentiment features in total, we do not test every possible subset of features, as the number of subsets grows exponentially with the number of features. Instead, we determine the F1-score when each sentiment feature is included individually. Then the 5 predictors with the highest F1-score are selected and all possible combinations of those 5 are tested. \\
\indent The result of our feature selection procedure is a best performing subset of features for each feature category (see Table \ref{tab:features}).

\subsection{Baseline}
To be able to evaluate our models against a benchmark, we develop a simple baseline classifier. The chosen model chosen takes a TF-IDF matrix of uni- and bi-grams. Here, all words occurring in more than 95\% or less than 5\% of the documents are removed. The TF-IDF is both passed to a LR and a RF classifier \shortcite{pedregosa_scikit-learn_2011}.

\begin{table}[!h]
    \centering
    \begin{tabular}{l|l}
        1 &  Most probable topic per tweet \\ \hline
        2 &  Neutral VADER Sentiment \\ \hline
        3 &  Positive VADER Sentiment \\ \hline
        4 &  Compound VADER Sentiment \\ \hline
        5 &  Positive Sentiment Disagreement \\ \hline
        6 &  TF-IDF Vector \\ \hline
        7 &  POS-unigrams \\
    \end{tabular}
    \caption{The obtained features after described feature selection process.}
    \label{tab:features}
\end{table}

\subsection{Final model} 
To choose the final classifier for the selected features we trained a LR, a SVM and a RF classifier as these were proven successful in the literature. The classifier were tested using grid-search and 5-fold cross-validation (see Appendix \ref{sec:appA} for the specific parameters tested). The final model resulted in a RF classifier.

% \subsubsection{Topic modelling}
% To further include a notion of content in addition to lexical features, we compute topic probabilities per document. Using latent dirichlet allocation (LDA), we find a suitable amount of topics based on perplexity measure. In this case, the chosen amount of topics is \textbf{WHAT???}.   

% \subsubsection{User embeddings}
% Following the definition of irony we expect ironic users to express something literally which is contradicted by the context of the users tweet history. Following \citeA{amir_modelling_2016} and \citeA{amir_quantifying_2017}, we compute user-embeddings to capture this. These are made by approximating the probability of each sentence per user and using negative samples (i.e words not rarely used by the user) to push the user further from those words and closer to commonly used words. Thereby the user is positioned in an embedding space based on its tweet history, perhaps close to similar users, encoding homophily across users. 

% In this paper, the user-embeddings are found using the USER2VEC framework proposed by \citeA{amir_quantifying_2017}\footnote{Code can be found at \url{https://github.com/samiroid/usr2vec}}. It is using the glove embedded vectors and results in a 200-dimensional vector per user. 

\subsection{Model}
Finally, we combined all remaining extracted and selected features (Table \ref{tab:features}) and trained a \texttt{sklearn.ensemble. RandomForestClassifier} after selecting the optimal parameters over \texttt{sklearn.model\_selection.GridSearchCV} (both packages taken from \shortciteA{pedregosa_scikit-learn_2011}).

\section{Results}\label{sec:results}
In the following section, we will highlight test results obtained from the different training procedures described in the method section. Baseline performance is shown in Table \ref{tab:BaselineLog} and \ref{tab:Baseline}. The best performing model from feature selection is shown in \ref{tab:FinalModel}. 

\begin{table}[h]
        \begin{subtable}[t]{0.5\textwidth}
        \centering
        %\rowcolors{1}{}{lightgray}
        \begin{tabular}{c|c}
        \textbf{Set} & \textbf{F1 score} \\
        \hline
        Train & 84,4\% \\ \hline
        Test & 60\%  
        \end{tabular}
        \caption{Baseline performance for LR classifier.}
        \label{tab:BaselineLog}
    \end{subtable}
    \begin{subtable}[t]{0.5\textwidth}
        \centering
        %\rowcolors{1}{}{lightgray}
        \begin{tabular}{c|c}
        \textbf{Set} & \textbf{F1 score} \\
        \hline
        Train & 85,1\% \\ \hline
        Test & 81\%  
        \end{tabular}
        \caption{Baseline performance for RF classifier.}
        \label{tab:Baseline}
    \end{subtable}
    \begin{subtable}[t]{0.5\textwidth}
        \centering
        %\rowcolors{1}{}{lightgray}
        \begin{tabular}{c|c}
        \textbf{Set} & \textbf{F1 score} \\
        \hline
        Train & 88.3\% \\ \hline
        Test & 84\%  
        \end{tabular}
        \caption{Final model performance with selected features.}
        \label{tab:FinalModel}    
    \end{subtable}
    \caption{Performance across candidate models. The performance for the train set is based on the mean F1 score over 5-fold cross validation.}
    \label{tab:performances_models}
\end{table}

\subsection{Baseline}

For the baseline we get an F1-score of 0.60 with a LR classifier; if we run it with RF, we see a drastic increase and reach a score of 0.81.

\subsection{Final model}
\begin{figure}[htb!]
    \centering
    \includegraphics[width=7.5cm]{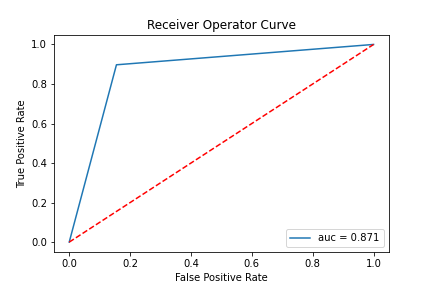}
    \caption{Receiver operator curve of the final model on the test set, where the Area Under the Curve (AUC) is indicated in the legend.}
    \label{fig:roc}
\end{figure}

The final model reaches an F1-score of 0.84 and has an Area Under the Receiver Operating Characteristics curve of 0.87 (see Figure \ref{fig:roc}) using the following features: a vector with the most probable topic per tweet, neutral, positive and compound sentiment vector with VADER, positive disagreement vector between RoBERTa and VADER, TF-IDF vector and POS-unigrams (see Table \ref{tab:features}). This is a substantial increase compared to the baseline results in seen in Table \ref{tab:BaselineLog}, and a small improvement in comparison to the RF baseline (Table \ref{tab:Baseline}).

For a more in-depth understanding of the model's behaviour, a confusion matrix is provided in Figure \ref{fig:Confusion}. These results show that the majority from each class are predicted correctly, with a slightly better prediction recall for irony (86 \%) in comparison to non-irony (82.35 \%). It also demonstrates a higher occurrence of false-positives (FP), i.e. non-ironic users predicted as ironic (17.65 \% FP). 

%%%% Confusion Matrix Body %%%%
\newcommand\MyBox[1]{%
    \fbox{\parbox[c][3cm][c]{3cm}{\centering #1}}%
    % Size of boxes
}
\newcommand\MyVBox[1]{%
    \parbox[c][1cm][c]{1cm}{\centering\bfseries #1}%
}  
\newcommand\MyHBox[2][\dimexpr3cm\fboxsep\relax]{%
    \parbox[c][1cm][c]{#1}{\centering\bfseries #2}%
}  
\newcommand\MyTBox[4]{%
    \MyVBox{#1}
    \MyBox{#2}\hspace*{-\fboxrule}%
    \MyBox{#3}\par\vspace{-\fboxrule}%
}  
%%%%
\newcommand*\rot{\rotatebox{90}}
\begin{figure}
    \begin{center}
    {

        \offinterlineskip

        \raisebox{-5.6cm}[0pt][0pt]{
            \parbox[c][3pt][c]{0.8cm}{\hspace{-3.5cm}\rot{\textbf{Actual}}\\[12pt]}}\par

        \hspace*{0.8cm}\MyHBox[\dimexpr3.4cm+6\fboxsep\relax]{Predicted}\par

        \hspace*{0.8cm}\MyHBox[\dimexpr3.4cm+6\fboxsep\relax]{Irony}\MyHBox[\dimexpr3.4cm+6\fboxsep\relax]{Non-irony}\par

        \MyTBox{\rot{Irony}}{50\\(86.00\%)}{8\\(13.79\%)}

        \MyTBox{\rot{Non-irony}}{12\\(17.65\%)}{56\\(82.35\%)}

    }
    \end{center}
    \caption{Confusion matrix on predicted classes for final model. Total Entries: 116 (test set). Percentages are shown in parenthesis.}
    \label{fig:Confusion}
\end{figure}

\subsection{Feature importance}
With the large amount of features, it is important to explore their influence individually and within their feature group. Every feature importance is documented in the Appendix \ref{sec:appA}, and the most notable results will be highlighted here. F1-scores reported in this section are based on 5-fold cross validation. 

\textbf{Topic}. Within the topic group, predictors reach up to F1-score of 58.19\%. The vector containing the dominant topic per tweet seem to be the most important feature as well as the driving factor when combining the features.

\textbf{Sentiment}. The different levels of importance within sentiment vary in general more than topic importance. Here the lowest score is 38.31\%. The sentiment scores provided by VADER seem to be among the best sentiment predictors. For example, the combination of neutral and positive vader scores reaches and F1-score of 59.90\%. A general observation is that the features benefit from being combined. The combination of neutral, positive, and compound VADER sentiment along with the agreement on positive sentiment between VADER and RoBERTa reaches an F1-score of 62.33\%. In comparison, features like the agreement alone only reaches 50.61\%. 

\textbf{Lexical}. Among the three groups, the lexical features possess the highest importance both, individually as well as in combination with each other. The TF-IDF itself reaches F1-score of 85.10\% and POS-Unigrams and mean tweet length reach 86.16\% and 67.85\%, respectively. Together they perform at a F1-score of 88.94\%. 

\section{Discussion}\label{sec:discussion}
This project has so far attempted to identify promising approaches within irony detection. It has developed a model using multiple features within the area of topics, sentiment and lexical items. This included a thorough training and testing of all features and comparing it to a baseline resulting in a Random Forest Classifier with a performance improvement of three percentage points.

\indent This section provides an in-depth discussion of the performance of the promising features and techniques (from Section \ref{sec:related}) in relation to our results. 

The result of our model positions itself within the general performance of irony detection models, however, it is not comparable to SOTA performances. Even though the baseline already has a high performance, adding selected sentiment and topic features still yield a slight improvement.

\subsection{Sentiment}

\textbf{Sentiment disagreement.} 
Surprisingly and in contrast to the presented literature (see Section \ref{sec:related}), we were not able to find a high contribution of sentiment disagreement to the overall prediction. 

The squared distance between the sentiment scores estimated by two different analyzers was chosen with the goal of revealing the harder-to-predict sentiment by the discrepancy of the lexicon and rule-based VADER and the transformer-based RoBERTa model motivated by early linguistic research \shortcite{wilson_verbal_1992, giora_irony_1995, barbe_irony_1995, utsumi_verbal_2000}. This was not successful. We attribute this observation to our assumption that both of the used analyzers would come to different predictions because of their different architecture. If the sentiment of a sentence would be hard to predict, the two analyzers might error in a similar way thus resulting in a low sentiment disagreement. For instance, the tweet 

\begin{table}[!ht]
    \centering
    \begin{tabular}{lP{6.5cm}}
       (1)  &  \textit{\#user\# goodness. the co wasn’t allowed to quietly retire? but wait … will orange julius intervene, like with that seal?}
    \end{tabular}
    \label{tab:tweet_example}
\end{table}

was classified with a negative, neutral, and positive VADER score as 0, 0.75, 0.25 and similar RoBERTa values of 0.059, 0.59, 0.36 resulting in very low disagreement scores. However, the label of the tweet's user is "ironic" which henceforth contrasts the prediction. At that point we suggest a further investigation of a larger number of different sentiment analyzers for future research.

\textbf{Sentiment contrast and variance.} 
We experience that the contrast and variance features captured with RoBERTa do not contribute a lot to the overall performance, neither as single nor in combination with other features. A possible reason could be that usually, RoBERTa performs best on entire sentences. \shortcite{liu_roberta_2019, wolf_transformers_2020}. As we used tri-grams to capture the contrast and variance within a sentence and especially its subsentences, RoBERTa might not have been sufficient at that point. It is, however, known to be robust and precise, for which it was originally chosen for the sentiment contrast task \cite{liu_roberta_2019}. 

The regular sentiment score provided by VADER remains as a solid feature which better captures irony than the features trying to capture the linguistic characteristics of irony. 

\subsection{Topic Modeling}

Against our expectation and not aligned with \shortciteA{nozza_unsupervised_2016}, topic models could not contribute very well to our models overall performance. Possible reasons are that i) despite our tests for the ideal number of topics, it still did not reflect the true number of topics but rather leaving room for ironic and non-ironic topics intersecting, and ii) our dataset is not separable into different topics per ironic/non-ironic label. This is, we see that most users across the dataset and both labels fall into the first topic.

Furthermore, we think that a more fine-grained topic analysis in combination with the sentiment could be beneficial. This follow the intuition that, if a human judge wasn't sure if a certain author tweets about a topic in an ironic way or not, they could take a look at the same-topic tweets in the user's history and identify their sentiment. This is, if a user would mainly tweet positively about a certain topic, the probability would be high that the underlying tweet is an ironic one if it is negative but about the same topic.

Lastly, we would like to suggest in a self-critical manner that training the model on the combined tweets per user rather than getting a topic score per tweet and averaging this over the respective user could show better performance. The reasons for this are, firstly, that it would more align with the other techniques we used and therefore make the data points better processable. Secondly, this would capture the actual task of an author classification better.

Despite the criticism of the topic model features, it is still worth mentioning that it can still play a role in the final model. It can be that it is a feature only providing little to some information on its own but contributes to a larger extent when combined with other features.  

\subsection{Lexical Features}
Generally, lexical features were the best performing features for labelling ironic users. POS unigrams performed best with an F1-score of 86.2\%, followed by TF-IDF (F1 of 85.1\%) and mean tweet length (F1 of 67.9\%). 

\begin{figure}[!h]
    \centering
    \includegraphics[width=7.5cm]{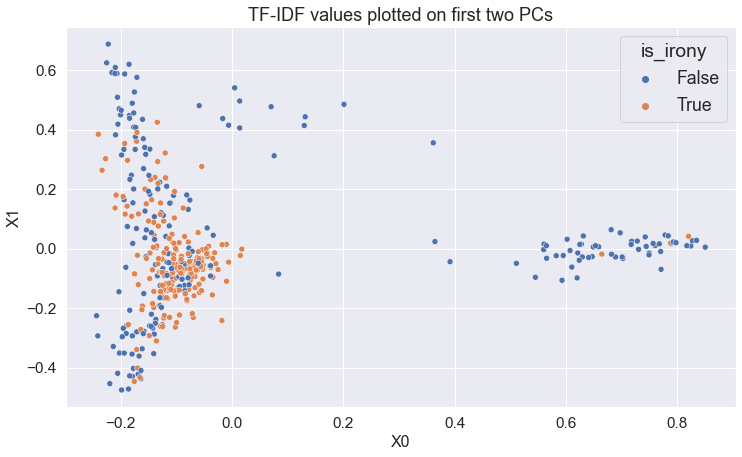}
    \caption{TF-IDF bigrams projected onto their first 2 PCs and colored in respective labels.}
    \label{fig:tfidf}
\end{figure}

Aligning with the literature \shortcite{khalifa_ensemble_2019}, we were able to achieve very good results for our TF-IDF analysis, already obtained at baseline level. In Table \ref{tab:most_freq_words}, we see that ironic and non-ironic authors have clearly different words. Frequent ironic uni- and bigrams have a "commenting" tone such as \textit{looks like} and \textit{calm down} whereas non-ironic terms are more related to society and communities (e.g. \textit{trans people}, \textit{military}\footnote{This is also interesting and contrasting in regards to the shared task's stereotyping aspect that addresses how minority groups are being stereotyped by using ironic language \shortcite{ortega-bueno_pan_2022}. However, this sample size of non-ironic bi-grams is too small to make meaningful predictions.}). Ironic users seem to be characterized by \textit{how} they talk and not \textit{what} they talk about. 
    A plot of the two principal components of the TF-IDF values confirm the importance of it to some extent (Figure \ref{fig:tfidf}). 

Here, we primarily observe two clusters; a cluster to the further right mostly consisting of non-ironic users and a larger cluster to the left of ironic users where non-ironic datapoints are scattered around it. Some non-ironic users seem easy to identify which might be a reason for the high performance and others are similar to ironic users which the confusion matrix (Figure \ref{fig:Confusion}) also confirms as there is a tendency to mistake non-ironic users as ironic. It is, however, evident that the frequent use of specific terms is a more explicit difference between ironic and non-ironic user than for example sentiment contrast or the topics dealt with.

\subsection{Further Work}
\textbf{Context.} As irony detection has seen great results when including context \shortcite{rangel_overview_2013, buschmeier_impact_2014}, we would suggest that further analysis should embrace a wider look into the meta-data of users. We think, that non-linguistic information like age and gender could potentially contribute to a better model as seen in \shortciteA{buschmeier_impact_2014}. Another way to integrate context is through user embeddings which have shown promise in detecting e.g. mental health and sarcasm detection on social media. In \shortciteA{amir_modelling_2016} and \shortciteA{amir_quantifying_2017}, such embedding is proposed. It is made by approximating the probability of each sentence per user and using negative samples (i.e. words rarely used by the user) to push the user further from those words and closer to commonly used words. Thereby the user is positioned in an embedding space based on its tweet history.  

\textbf{Content.} As previously emphasized, topic modelling is an understudied feature in relation to irony and perhaps it is needed to start by replicating \citeauthor{nozza_unsupervised_2016}'s \citeyear{nozza_unsupervised_2016} results with the dataset used in this project. Additionally, further investigations could for example include a wider range of other topic modeling approaches.
Furthermore, we believe that also the more fine-grained look into the masked items \textit{\#HASHTAG\#}, \textit{\#USER\#} and \textit{\#URL\#} could give a more detailed possibility of analysing the users intentions. We suggest that one could combine the research methods presented in this paper with an investigation of the specific hastags, url, and retweets.

\textbf{Robustness.} Currently, our model only works for predicting whether an previously unseen user is ironic or not if the model receives the features (as in Table \ref{tab:features}) of 200 tweets as input for this user. For future work, it would be interesting to work towards an approach that can take a varying number of tweets/user for making predictions about unseen users. A way this could be achieved, is by concatenating all tweets into one document per user, and then extracting the features on a user level based on that. Another way could be to extract the features per tweet, but then calculating the mean of each feature per user before feeding it into the model. 

It seems like there is a difficulty in capturing irony through its sentimental aspects (such as contrast). The ambition to actually capture irony by its linguistic characteristics has proven to be difficult. Furthermore, topic seems to contribute quite little in isolation, suggesting that in terms of irony, we cannot determine whether a users is ironic based on the topic or content itself. The ironic users may simply talk about all sorts of topics ironically. Lastly, the TF-IDF results emphasize that the most common words and bi-grams for irony are not related to specific content as such but rather the way users talk about content.

%The standard deviation of sentiment scores has been proven to be a well performing factor for our model.

\section{Conclusion}\label{sec:conclusion}
This study proposed a model for detecting figurative language and more precisely, irony, on the social media platform Twitter. Based on a thorough literature review, the chosen features for the task extracts content of the tweets such as topics and sentiment contrasts, and lexical features. An extensive feature- and model-selection process led to a final model performing at an F1-score of 0.84, positioning the model within the general performance of language models detecting irony. The feature selection process revealed a large benefit of including lexical features such as TF-IDF in the models. Ironic and non-ironic users simply seem to use different terms more frequently instead of portraying contrasting sentiments or very specific topics. Further work should investigate the importance of topic models and how to incorporate the context of each user, for example through user embeddings as well as use the content of a hashtag, retweets and urls.  

\bibliographystyle{apacite}
\bibliography{langprocapi}

\appendix
\clearpage
\onecolumn

\section{Results of Feature Selection for Each Group}\label{sec:appA}

\begin{table}[!ht]
    \footnotesize
    \centering
    \begin{tabular}{l|c}
    \toprule
    Predictors &       F1 \\
    \midrule
    'pos\_sent\_vecs' & 0.395431 \\
    'neg\_sent\_vecs' & 0.437334 \\
    'X\_negative' & 0.438687 \\
    'X\_neutral' & 0.546429 \\
    'X\_positive' & 0.437568 \\
    'negVader' & 0.474704 \\
    'neuVader' & 0.562881 \\
    'posVader' & 0.539276 \\
    'compoundVader' & 0.538492 \\
    'diff\_neg' & 0.475638 \\
    'diff\_pos' & 0.506140 \\
    'diff\_neu' & 0.427438 \\
    'pos\_sent\_std' & 0.401262 \\
    'neg\_sent\_std' & 0.383127 \\
    'X\_neutral', 'neuVader' & 0.512140 \\
    'X\_neutral', 'posVader' & 0.560650 \\
    'X\_neutral', 'compoundVader' & 0.543192 \\
    'X\_neutral', 'diff\_pos' & 0.478161 \\
    'neuVader', 'posVader' & 0.599900 \\
    'neuVader', 'compoundVader' & 0.548793 \\
    'neuVader', 'diff\_pos' & 0.550010 \\
    'posVader', 'compoundVader' & 0.588643 \\
    'posVader', 'diff\_pos' & 0.562398 \\
    'compoundVader', 'diff\_pos' & 0.545492 \\
    'X\_neutral', 'neuVader', 'posVader' & 0.504389 \\
    'X\_neutral', 'neuVader', 'compoundVader' & 0.572260 \\
    'X\_neutral', 'neuVader', 'diff\_pos' & 0.573182 \\
    'X\_neutral', 'posVader', 'compoundVader' & 0.534362 \\
    'X\_neutral', 'posVader', 'diff\_pos' & 0.455556 \\
    'X\_neutral', 'compoundVader', 'diff\_pos' & 0.477582 \\
    'neuVader', 'posVader', 'compoundVader' & 0.567817 \\
    'neuVader', 'posVader', 'diff\_pos' & 0.524117 \\
    'neuVader', 'compoundVader', 'diff\_pos' & 0.539629 \\
    'posVader', 'compoundVader', 'diff\_pos' & 0.521223 \\
    'X\_neutral', 'neuVader', 'posVader', 'compoundVader' & 0.605726 \\
    'X\_neutral', 'neuVader', 'posVader', 'diff\_pos' & 0.551243 \\
    'X\_neutral', 'neuVader', 'compoundVader', 'diff\_pos' & 0.588719 \\
    'X\_neutral', 'posVader', 'compoundVader', 'diff\_pos' & 0.573983 \\
    'neuVader', 'posVader', 'compoundVader', 'diff\_pos' & 0.623307 \\
    'X\_neutral', 'neuVader', 'posVader', 'compoundVader', 'diff\_pos' & 0.608025 \\
    \bottomrule
    \end{tabular}
\label{tab:sentiment}
\caption{All feature performances within the sentiment group.}
\end{table}

\begin{table}[h]
    \footnotesize
    \centering
    \begin{tabular}{l|c}
    \toprule
        Predictors & F1 \\
    \midrule
        'dominant\_topic\_user' & 0.00 \\
        'cluster' & 0.53 \\
        'max\_probabilities' & 0.58 \\
        'max\_probabilities', 'dominant\_topic\_user' & 0.58 \\
        'max\_probabilities', 'cluster' & 0.58 \\
        'dominant\_topic\_user', 'cluster' & 0.00 \\
        'max\_probabilities', 'dominant\_topic\_user', 'cluster' & 0.58 \\
    \bottomrule
    \end{tabular}
\label{tab:topic}
\caption{Feature performances within the topic group.}
\end{table}

\begin{table}[h]
    \footnotesize
    \centering
    \begin{tabular}{l|c}
    \toprule
    Predictors &       F1 \\
    \midrule
    'tfidf' & 0.850981 \\
    'pos\_unis' & 0.861648 \\
    'mean\_len' & 0.678526 \\
    'tfidf', 'pos\_unis' & 0.889478 \\
    'tfidf', 'mean\_len' & 0.850981 \\
    'pos\_unis', 'mean\_len' & 0.861648 \\
    'tfidf', 'pos\_unis', 'mean\_len' & 0.889478 \\
    \bottomrule
    \end{tabular}
\label{tab:sentiment}
\caption{Feature performances within the lexical group.}
\end{table}

\begin{figure}[!ht]
    \footnotesize
    \centering
    \includegraphics[width=14cm]{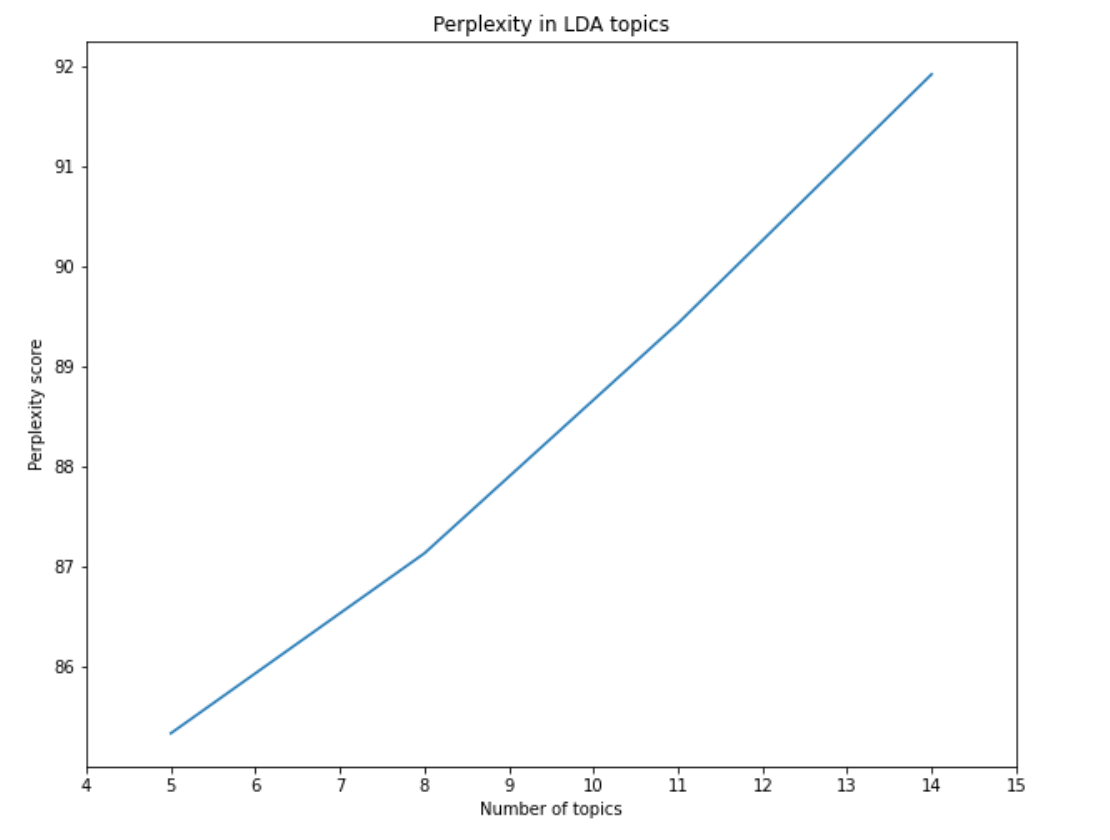}
    \caption{Perplexity scores when testing values between 5 and 14. The lowest perplexity score is found at 5 topics with a score of 85.33.}
    \label{fig:perplexity}
\end{figure}

\vfill
\section{Hyper-parameters}\label{sec:appB}
The following parameters were tested with five-fold gridsearch:
\vfill
\textbf{Logistic regression:}
\begin{itemize}
    \item Penalties: L2
    \item C: [1.0, 0.1, 0.01] 
    \item Solvers: ['liblinear', 'sag', 'saga', 'lbfgs']
\end{itemize}

\textbf{Random Forest:}
\begin{itemize}
    \item Number of estimators: [200, 500]
    \item Max features: ['auto', 'sqrt']
    \item Maximum depth: [4,5,6,7,8]
    \item Criterion:['gini', 'entropy']
\end{itemize}

\textbf{Support vector machine:}
\begin{itemize}
    \item Kernel: ['linear', 'rbf', 'sigmoid']
    \item C: [1, 2, 3]
\end{itemize}

\label{sec:appendix}

\end{document}